\documentclass[final]{cvpr}

\usepackage{times}
\usepackage{epsfig}
\usepackage{graphicx}
\usepackage{amsmath}
\usepackage{amssymb}
\usepackage{pifont}
\usepackage{diagbox}
\usepackage[pagebackref=true,breaklinks=true,colorlinks,bookmarks=false]{hyperref}

\def\cvprPaperID{2514} 
\def\confYear{CVPR 2021}

\newcommand\blfootnote[1]{
  \begingroup
  \renewcommand\thefootnote{}\footnote{#1}
  \addtocounter{footnote}{-1}
  \endgroup
}

\begin{document}

\title{Multimodal Motion Prediction with Stacked Transformers}

\author{
   {Yicheng Liu}${^{1\star}}$ \quad
   {Jinghuai Zhang}${^{2\star}}$ \quad
   {Liangji Fang}${^{2}}$ \quad 
   {Qinhong Jiang}${^{2}}$ \quad
   {Bolei Zhou}${^{1}}$\\
   {The Chinese University of Hong Kong}${^1}$\qquad
   {SenseTime Research}${^2}$ \qquad \\
}

\maketitle

\blfootnote{$^{\star}$ Co-first authors with equal contributions.}

\begin{abstract}
Predicting multiple plausible future trajectories of the nearby vehicles is crucial for the safety of autonomous driving.
Recent motion prediction approaches attempt to achieve such multimodal motion prediction by implicitly regularizing the feature or explicitly generating multiple candidate proposals. 
However, it remains challenging since the latent features may concentrate on the most frequent mode of the data while the proposal-based methods depend largely on the prior knowledge to generate and select the proposals.  
In this work, we propose a novel transformer framework for multimodal motion prediction, termed as mmTransformer.
A novel network architecture based on stacked transformers is designed to model the multimodality at feature level with a set of fixed independent proposals. 
A region-based training strategy is then developed to induce the multimodality of the generated proposals.
Experiments on Argoverse dataset show that the proposed model achieves the state-of-the-art performance on motion prediction, substantially improving the diversity and the accuracy of the predicted trajectories. Demo video and code are available at  \url{https://decisionforce.github.io/mmTransformer}.

\end{abstract}

\section{Introduction}
Predicting the future trajectories of nearby vehicles is critical for the Autonomous Vehicle systems to understand the surrounding and make informative decisions. Multimodal prediction, which aims to generate multiple plausible trajectories of the target vehicle, plays a key role to handle the uncertainty in motion prediction and improve the safety of motion planning. 
\begin{figure}
    \centering
    \includegraphics[width =0.9\linewidth]{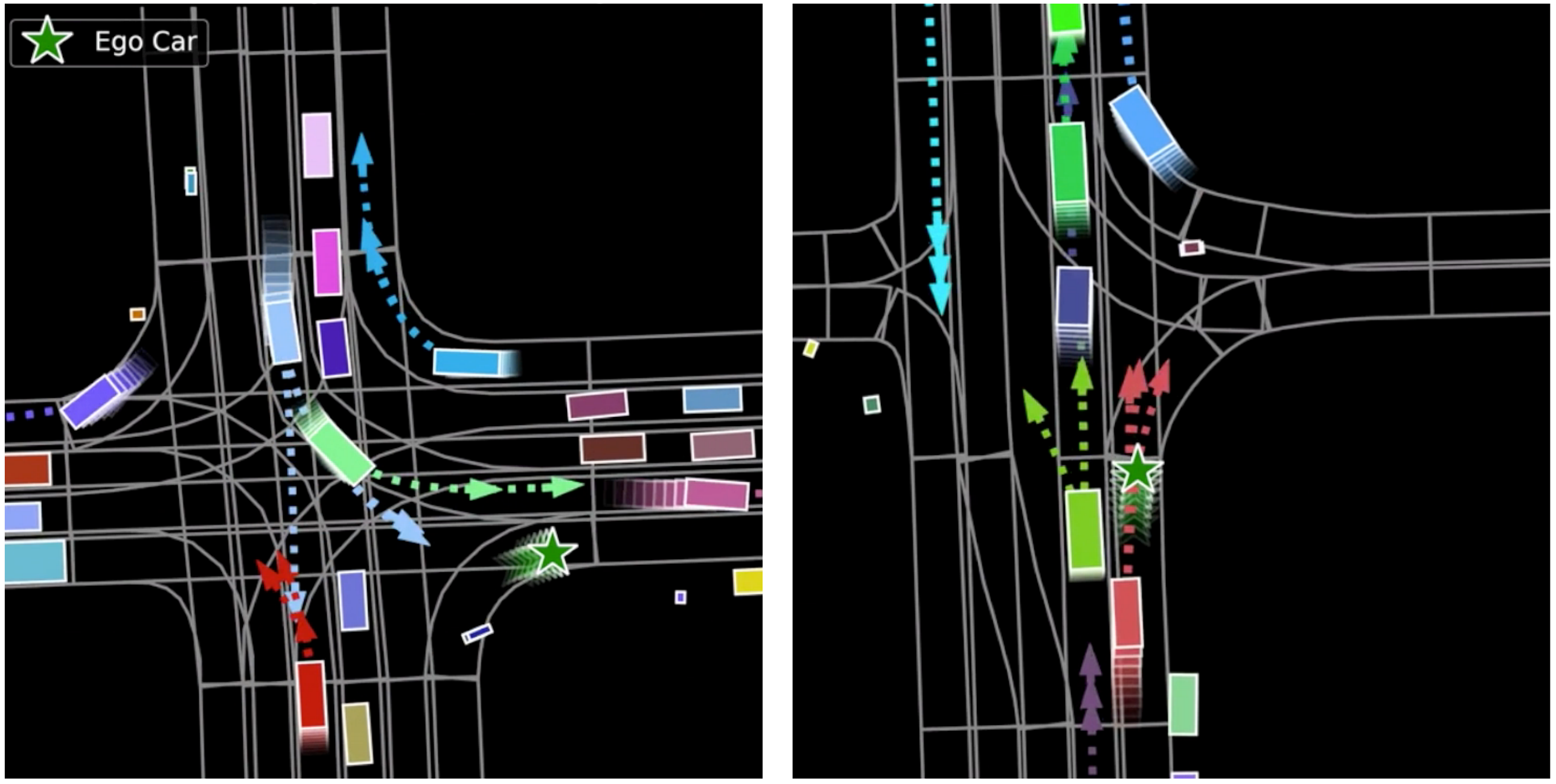}
    \caption{Examples of multimodal motion prediction in complex driving scenarios. For each moving vehicle near the ego car, three plausible future trajectories are predicted by the proposed model.}
    \label{fig:Teaser}
\end{figure}
Due to the uncertain future events, traffic vehicles could perform differently even under the same scene. However, there is only one ground truth trajectory collected in each driving scene.
Hence one challenge for enabling multimodal prediction lies in how to learn to cover all the possible outcomes in a given scene with limited training samples.

Recent motion prediction methods mainly follow probabilistic approaches \cite{lee2017desire,hong2019rules,tang2019multiple} or proposal-based approaches \cite{zhao2020tnt, phan2020covernet,chai2019multipath, fang2020tpnet} to address the aforementioned issue.
The probabilistic approaches implicitly model the uncertainty of the trajectory through defining the underlying possible models as a latent variable. They either achieve the multimodal prediction with generator conditioned on different latent variables, or directly constrain the output over a probability distribution(\eg, GMM) to get diverse results. 
These methods depend heavily on the predefined prior distribution and the well-designed loss function, which might be prone to the optimization instability and the mode collapse issue.
Unlike probabilistic approaches which generate multimodal outputs through modeling the latent distribution of the modality, the proposal-based approaches \cite{fang2020tpnet,zhao2020tnt,chai2019multipath,pan2019lane} perform in an alternative way, which first defines candidate points or trajectories as proposals, and then regress or classify these proposals to the ground truth. With predefined proposals, these methods alleviate optimization burden and narrow down the feasible space of solutions. Although these methods achieve good performance, they still have the following two issues:
1) The result relies heavily on the quality of the predefined anchors since the heuristic methods are applied to sample the candidate points.
2) The multimodal prediction can not be guaranteed since multimodal nature of trajectory prediction is not well captured with only one ground truth provided during the training.

In this work, we propose a novel end-to-end multimodal motion prediction framework called MultiModal Transformer (\textbf{mmTransformer}), where the proposals are first randomly initialized and then refined to incorporate contextual information.
mmTransformer is designed based on the transformer architecture, which proves to be effective in modeling sequential data.
The whole model can be viewed as stacked transformers in which the past trajectories, the road information, and the social interaction are aggregated hierarchically with several transformer encoder-decoder modules. Two multimodal prediction examples of the whole traffic scenes are shown in Fig~\ref{fig:Teaser}.

We develop two new mechanisms to ameliorate the uni-modal effects brought by identical features.
First, we introduce a trajectory proposal mechanism to the field of motion prediction. Specifically, queries in the decoders of mmTransformer are represented as trajectory proposals, which asymptotically aggregate multiple channels of contextual information from encoders, and make independent predictions. Since these proposals are orthogonal with each other, each of them will carry customized features, which promotes the diversity and multimodality.
Second, a region-based training strategy (RTS) is developed to explicitly ensure the multimodality, which negotiates the conflicts between the uniqueness of ground truth and multimodal nature of predictions. We divide the surrounding space into several regions and group trajectory proposals into different sets, with each set being assigned to one region. During training, only the set of proposals assigned to the region where ground truth locates will be utilized to optimize the framework.
This new strategy enforces individual proposal to focus on a specific mode, without compromising the latent features learned by other proposals.

The contributions of this paper are summarized as follows: (1) To the best of our knowledge, mmTransformer is the first model using stacked transformers for trajectory proposals to aggregate multiple channels of contextual information and achieve multimodal prediction.
(2) To preserve the multimodal nature of motion forecasting, we design a novel region-based training strategy, which ensures that each individual proposal is capable of capturing a specific mode. 
(3) Extensive experiments show the substantial improvement brought by the proposed model architecture and the tailored region-based training strategy. Our model ranked the 1st on the Leaderboard of Argoverse benchmark dated on 16 Nov 2020, and remains competitive on the leaderboard.

\begin{figure*}
    \centering
    \includegraphics[width =0.9\textwidth]{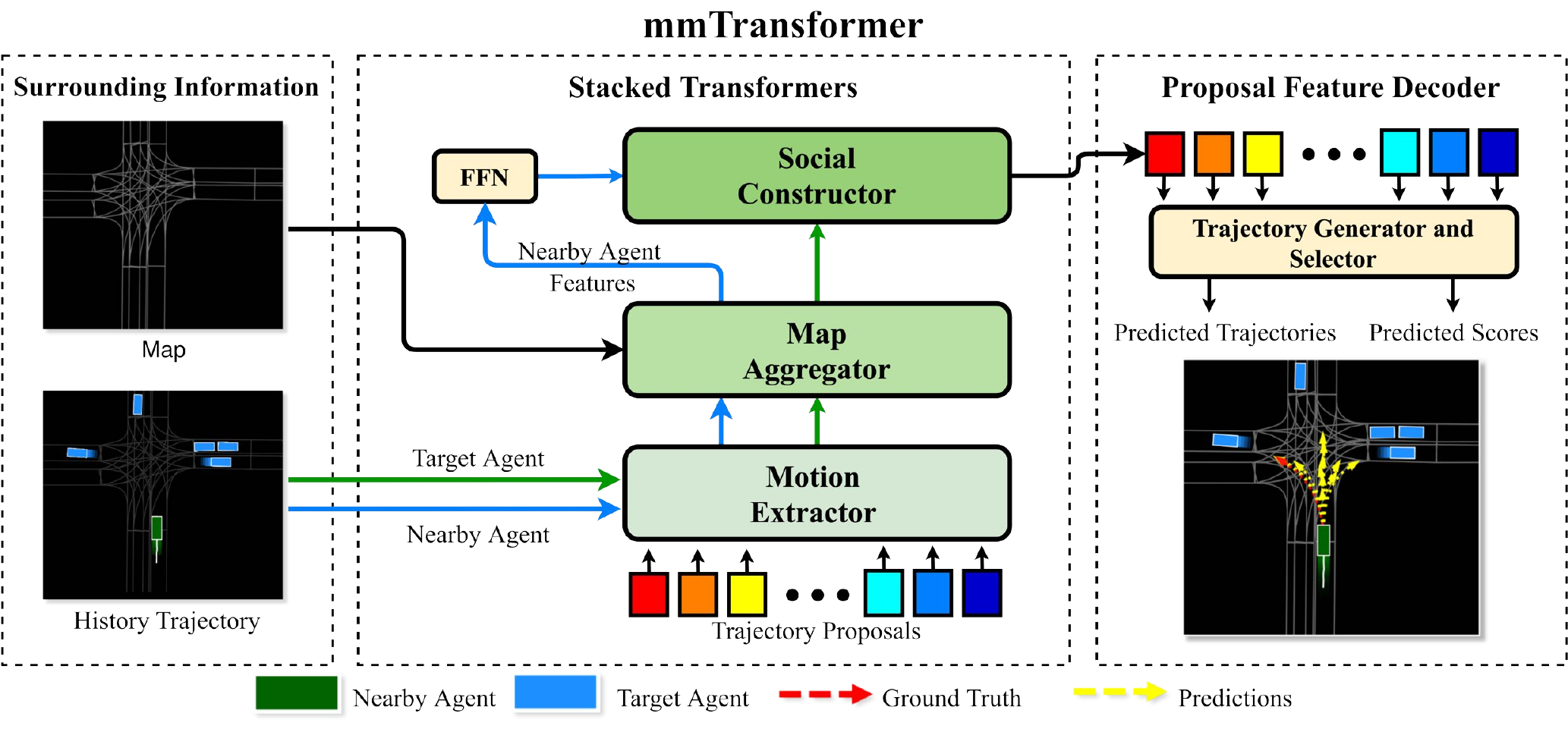}
    \caption{\textbf{Overview of mmTransformer:} The proposed mmTransformer uses stacked transformers as the backbone, which aggregates the contextual information on the fixed trajectory proposals. 
    Proposal feature decoder further generates the trajectories and corresponding confidence scores by decoding each proposal feature through the trajectory generator and selector, respectively. Besides, the trajectory generator and trajectory selector are two feed-forward network, which have the same structure with the FFN in the transformer \cite{vaswani2017attention}. The detailed network architecture can be found in appendix.}
    \label{fig:frame work}
\end{figure*}

\section{Related Work}
\label{sec:related}

\paragraph{Motion Prediction.}
Recently deep learning-based methods have been widely used for motion prediction \cite{alahi2016social, sgan, deo2018convolutional,altche2017lstm, fang2020tpnet}. The typical pipeline is to first create the input representation by rasterizing \cite{djuric2018motion, zhang2020novel} or vectorizing \cite{chu2019neural,liang2020learning,gao2020vectornet} surrounding information and then use deep neural network (\eg, CNN, Long Short-Term Memory(LSTM), graph neural network\cite{yagi2018future,hochreiter1997long}) to extract informative features. Finally, the trajectory is directly generated by model \cite{gao2020vectornet} or based on prior knowledge \cite{fang2020tpnet, zhao2020tnt, chai2019multipath}.

To this end, motion prediction methods can be roughly divided into two categories: feature-based and proposal-based methods. For the first kind of methods, most of them focus on how to extract useful information from the environment. CNN Encoder-Decoder is proposed in \cite{yagi2018future} to extract features from vehicles’ past positions and directions and directly regress the future positions. Graph neural networks \cite{liang2020learning} have emerged in response to problems that scenes cannot be easily represented by matrices of pixels or simple vectors. SoPhie \cite{sadeghian2019sophie} have leveraged features extracted from the physical environment, attention mechanisms, and adversarial training.
However, for these feature-based methods it is difficult to guarantee the multimodal prediction of the model. As for another kinds of models \cite{zhao2020tnt, phan2020covernet,chai2019multipath, fang2020tpnet}, where the candidate trajectory set is first generated based on prior knowledge and candidate point and then optimize and reduce these candidate trajectory by designed cost function or post-processing  to obtain final prediction. Although these methods successfully modeling the multimodality, there are still many unsolved problems.
As these methods focus on the manipulation of the predicted trajectories or candidates, the meanings of latent features are typically neglected, which may hurts the stability of the model. As a result, performance will be significantly influenced by the robustness of prior hypothesis made by the author.

Different from above, our model achieves the multimodal prediction at both feature and proposal levels. The region-based training strategy further refines the proposals which reduces the correlation of the proposals and guarantees the diversity of the predicted trajectories. 
Meanwhile, by explicitly considering modality into proposal during training process, mmTransformer has a more interpretable pipeline.

\paragraph{Transformer.}
Transformer is a novel attention-based method which was firstly introduced in \cite{vaswani2017attention}. It has been successfully deployed in several applications (\eg, neural machine translation and image caption generation \cite{devlin2018bert, luscher2019rwth}). The most important part of transformer is the self-attention mechanism.
The advantage of the attention mechanism in transformer lies in its capability of learning high quality features through taking the whole context into consideration.
Some of the recent methods in the field of trajectory forecasting adopt the attention mechanisms in sequence and interaction modeling \cite{li2020end,giuliari2020transformer,mercat2020multi}. For example, an interaction transformer \cite{li2020end} is introduced to model the interaction between traffic vehicles. Ind-TF \cite{giuliari2020transformer} replaces RNN with vanilla transformer to model the trajectory sequences. 
Unlike these methods that use transformer as a part of their feature extractor, a fully transformer based architecture is used in our case to solve the multimodal motion prediction problem.

\section{Multimodal Motion Prediction Framework}
\label{sec:model}

Motion prediction aims to accurately predict the future motion of target vehicles, given the history trajectories of traffic vehicles in the scene and other contextual information such as road and traffic information. 
To tackle the multimodal motion prediction, we firstly learn a feature set $\mathcal{Y}$ comprising various proposal features $\boldsymbol{y} \in \mathcal{Y}$. Each $\boldsymbol{y}$ is generated from $\mathcal{F}_{\theta}(\boldsymbol{x})$, where $\boldsymbol{x}$ is the scene information, involving motion history and surrounding context. With the set of proposal features, we can generate multiple future trajectories $\rm S =  \{ \boldsymbol{s}_i \in \mathbb{R}^{T \times 2}: 1 \leq i \leq K \}$ by $\mathcal{G}_{\phi}(\boldsymbol{y})$, in which $T$ denotes future horizon and $K$ denotes the total number of predictions. Additionally, prediction set $S \subset \mathcal{S}$ where $\mathcal{S}$ is the entire space of possible $S$, and $\mathcal{F}_{\theta}(\cdot)$ and $\mathcal{G}_{\phi}(\cdot)$ are parameterized by $\theta$ and $\phi$, respectively. 

We aim to construct an appropriate set of proposal features $Y \subset \mathcal{Y}$ to ensure both accuracy and multimodality. Therefore, we introduce a novel mmTransformer to obtain informative proposals and then apply the region-based training strategy to ensure the multimodality of proposals.

\subsection{Stacked Transformers}
Transformer has demonstrated outstanding performance in dealing with sequential data. In order to apply transformer to trajectory prediction, we need to extend the model to incorporate a variety of the contextual information, because the vanilla transformer only supports encoding single type of data (\eg, the corpus token in the language transformer \cite{devlin2018bert}, and image in the visual transformer \cite{carion2020end}). 
A naive solution is to concatenate all types of inputs such as past trajectory and lane information into a sequence of contextual embeddings and input them to the transformer. As the transformer requires a fixed size of the input, a naive solution will consume a large amount of resources. Additionally, since different types of information will compound in such design and be aggregated by the attention layers, the quality of the latent feature might be compromised. Therefore, we consider the alternative of incorporating multiple channels of information separately.

Under the circumstance of different inputs, the challenge lies in how to incorporate multiple channels of information as input to the transformer. 
we propose to define the queries of transformer decoder as trajectory proposals, tailored to our multimodal trajectory prediction task. This design is inspired by the parallel version of transformer used in \cite{carion2020end}. 
Its strength is that parallel trajectory proposals can integrate the information from the encoder independently, allowing each single proposal to carry disentangled modality information. 
The stacked architecture adapts to the multi-input circumstance with several tailored feature extractors, integrating different contextual information hierarchically. 
Specifically, the structure of the stacked transformers consists of three individual transformer units, motion extractor, map aggregator, and social constructor respectively, each taking the updated trajectory proposals from the previous transformer as the input of its decoder to refine the proposals. The framework is illustrated in Fig~\ref{fig:frame work}.

For simplicity, the transformer modules retain the structure introduced by \cite{carion2020end}. Since the transformer decoder is permutation-invariant, the $K$ proposals must be distinct to each other in order to represent different modes and generate different trajectories. A learned positional encoding is added before each decoder layer of the transformer modules. We illustrate the three components in more detail as follows. 

\noindent\textbf{Motion Extractor.} 
The encoder input of motion extractor is the history trajectories of observed vehicles as $H= \{ \boldsymbol{h}_{i}\in \mathbb{R}^{T_{obs}\times 2}: 1 \leq i \leq N_{vehicle} \}$, where $N_{vehicle}$ is the number of the observed vehicles, including the target vehicle, and $T_{obs}$ is the length of history observation. The decoder inputs are the trajectory proposals $Y=\{ \boldsymbol{y}_i \in \mathbb{R}^{n}: 1 \leq i \leq K\}$, textcolor{red}{which are initialized by a set of learnable positional encoding, feature size of each encoding is $n$}. The outputs of decoder can be considered as proposal features. It is noted that all the observed vehicles share the same architecture of motion extractor, the same for map aggregator.

\noindent\textbf{Map Aggregator.} 
As the behaviors of vehicles depend largely on the topology of the map, such as road structure, we utilize this transformer to fuse the geometric and semantic information from high definition map to refine features of the input proposals. Following \cite{gao2020vectornet}, we encode each centerline segment into vectorized representation, and use the subgraph module in \cite{gao2020vectornet} to process each vectorized polylines. After that, the latent features of the polylines are fed to the map aggregator to explicitly model the scene.
Benefiting from the interpretability of encoder-decoder attention module inside transformer, the proposal can retrieve the corresponding map features based on its preassigned modality. 

\noindent\textbf{Social Constructor.} 
Unlike the previous transformers, social constructor encodes the vehicle features among all observed vehicles, aiming to model the interactions between them. In particular, the vehicle feature of each observed vehicle is obtained by summarizing all the proposals for each of vehicles via a multi-layer perception (MLP). The vehicle feature can be also viewed as distribution of future movements for each traffic vehicle. 
Since our objective is to forecast the future trajectory of the target vehicle, we only utilize the decoder of social constructor to update the proposals for target vehicles, instead of all vehicles, in pursuit of higher efficiency.

As a whole, the motivation behind our framework is to establish the intra-relation inside data (\eg, extracting the map topology with the encoder of map aggregator), and integrate contextual information from different encoders asymptotically to update each proposal and highlight its pre-assigned modality. 

Intuitively, stacked transformers can be divided into two parts. First part only encodes the information of each vehicle individually by the motion extractor and the map aggregator, without any interaction information being involved. Then the social constructor is applied to aggregate the nearby information and model the dependency among all the vehicles. Thus, the order is logically sequential, \ie, the social relation should be constructed based on individual vehicle features.
Additionally, the order of the other transformers has been verified empirically by experimental results shown in Table~\ref{tab:important_of_each_part}. 
Based on this stacked architecture, our model can capture the latent connection between contextual information and the diverse proposal features which ensures the multimodal predictions.   

\subsection{Proposal Feature Decoder}
The final Proposal Feature Decoder comprises two branches, namely, the Trajectory Generator for trajectory prediction and the Trajectory Selector branch for proposal scoring. For each of the $K$ target proposals, we apply a three-layer MLP $\mathcal{G}(\cdot)$ to generate the prediction as follows,
\begin{equation}
    \rm
    S = \{\boldsymbol{s}_i\in \mathbb{R}^{T \times 2}: \boldsymbol{s}_i=\mathcal{G}(\boldsymbol{y}_i), \boldsymbol{y}_i \in Y, 1 \leq i \leq K \}
\end{equation}
where $y_i \in Y$ is the $i^{th}$ proposal feature generated from social constructor, $\boldsymbol{s}$ is a tensor of predicted trajectory, $T$ is the number of future time steps.

For scoring, we apply the MLP, $\mathcal{W}(\cdot)$, with the same structure as regression branch to generate the $K$ confidence scores for each of the trajectory proposals.

\begin{gather}
    \rm
    C = \{c_i \in \mathbb{R} : c_i = \mathcal{W}(\boldsymbol{y}_i), 1 \leq i \leq K,\} 
\end{gather}

\begin{figure}
    \centering
    \includegraphics[width=0.9\linewidth]{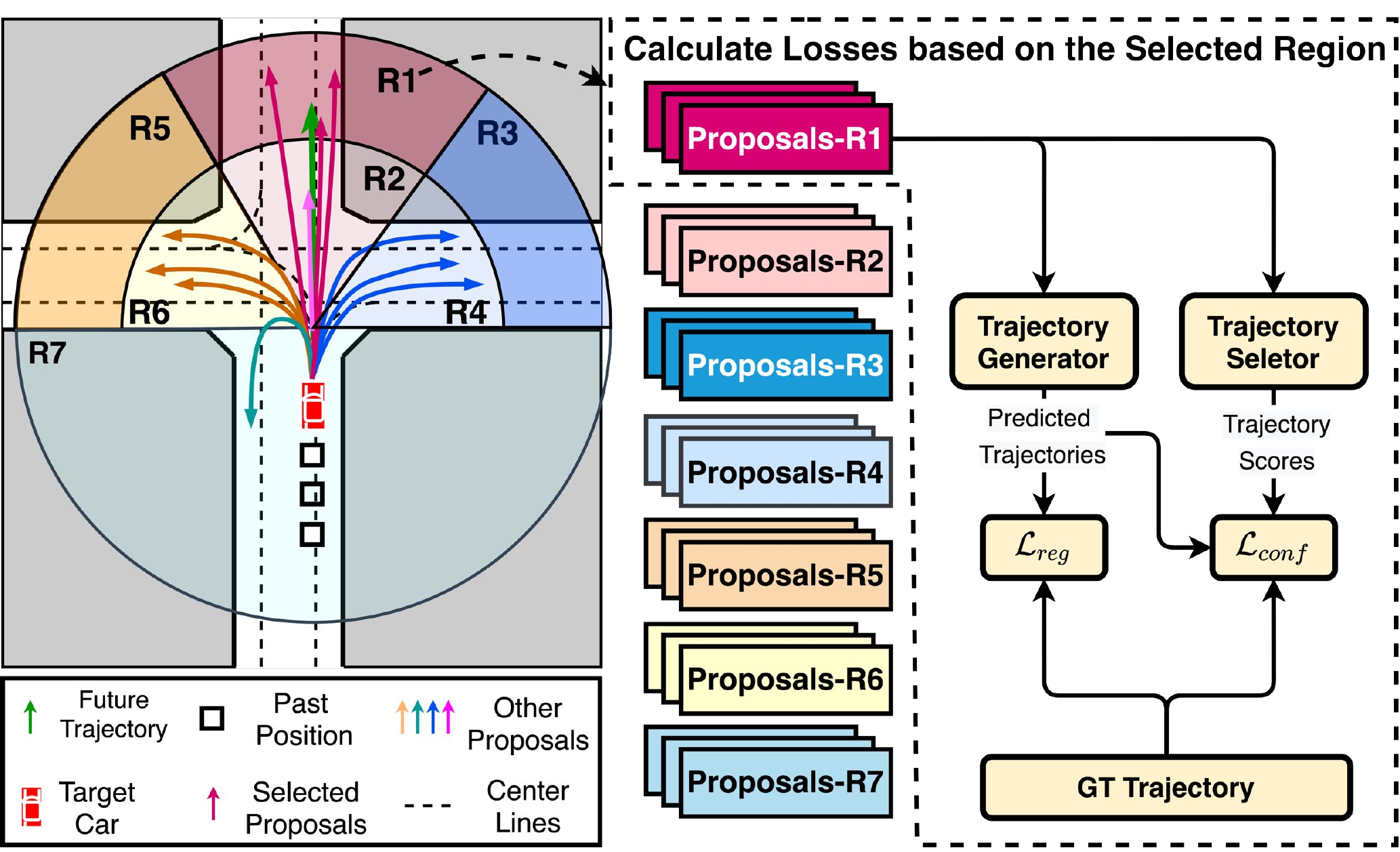}
    \caption{Overview of the region-based training strategy. We first distribute each proposal to one of the $M$ regions. These proposals, shown in colored rectangles, learn corresponding proposal feature through the mmTransformer. Then we select the proposals assigned to the region where the GT endpoints locate, generate their trajectories as well as their associated confidence scores, and calculate the losses for them.}
    \label{fig:region-based Training strategy}
\end{figure}
\subsection{Region-based Training Strategy}
As shown in previous work \cite{zhao2020tnt}, direct regression for all the trajectory proposals leads to the mode average problem, which hampers the multimodalily property of the model. One feasible solution to overcome the mode average problem is to calculate the regression loss and classification loss only using the proposal with the minimum final displacement error. We term this as the vanilla training strategy. Although our model achieves competitive results under this training strategy using 6 trajectory proposals, the prediction results are confined to a local region around the ground truth trajectory due to the small number of proposals are used. However, the modality collapsing issue occurs when we attempt to increase the number of proposals to improve diversity.

To address the limitation of the vanilla training strategy, we propose a novel training strategy called region-based training strategy (RTS), which groups trajectory proposals into several spatial clusters based on the spatial distribution of the ground truth endpoints, and optimizes the framework to improve the prediction results within each cluster.

The strategy is illustrated in Fig~\ref{fig:region-based Training strategy}. For each single scenario, we first rotate the scene to align the heading direction of the target vehicle to $+y$ axis, and make all coordinates centered at the last observation point of the target vehicle. Based on that, we partition the sample space of target vehicles into $M$ regions, without any overlaps between them. The detailed analysis of region shape and region number is illustrated in Section~\ref{sec:exp_xxx} and the procedure of the partition is illustrated in appendix. After that, we equally divide the total $K$ proposals of mmTransformer into $M$ parts, with each of them assigned to a specific region. As a result, each region will possess $N$ individual proposals, where $N=K/M$. It is noted that the pre-processing in our work ensures that all the samples can share the same partition map.

During the training, we utilize the regression loss and classification loss in a similar way to vanilla training strategy. The difference is that we calculate the loss for all the proposals that are assigned to the region where ground truth endpoint locate, rather than the one closest to the ground truth. In this way, we improve the multimodal results in a region-based manner, which optimizes the predictions in one region without affecting any other regions.

\subsection{Training Objective}
Since all the modules are differentiable, our framework yields fully supervised end-to-end training. The final loss of our model is composed of regression loss, scoring loss, and an auxiliary loss for multi-task learning. They are the Huber loss for regression, the KL divergence for scoring trajectories, and an auxiliary loss for the region classification respectively. The detail of losses is as follows:

\noindent\textbf{Regression Loss.} $\rm \mathcal{L}_{reg}$ is the Huber loss over per-step coordinate offsets.
\begin{equation}
    \rm
    \mathcal{L}_{reg} = \frac{1}{N} \sum_{i=1}^{N} \mathcal{L}_{Huber}(\boldsymbol{s}_i, \boldsymbol{s}_{gt}),
\end{equation}
where $\boldsymbol{s}^i$ is the $i$-th predicted trajectory generated by proposal feature decoder, and the $\boldsymbol{s}^{gt}$ is the ground truth trajectory. 

\noindent\textbf{Confidence Loss.}To assign each trajectory a confidence score, we follow the \cite{zhao2020tnt} to tackle this scoring problem of trajectory prediction via a maximum entropy model,

\begin{gather}
    \rm
    \tau(\boldsymbol{y}) = \frac{\exp{(\mathcal{W}(\boldsymbol{y}))}}{\sum_{i=1}^N \exp{(\mathcal{W}(\boldsymbol{y}_i))}}, \\
    Y_{region} = \{ \boldsymbol{y}_1, \dots, \boldsymbol{y}_N \}, \notag
\end{gather}
where $Y_{region}\subset Y $ is a subset of proposal features selected
by the region-based training strategy from $Y$.

\begin{equation}
    \rm
    \lambda(\boldsymbol{s}) = \frac{\exp{(-\mathcal{D}(\boldsymbol{s}, \boldsymbol{s}_{gt}))}}{\sum_{i=1}^N \exp{(-\mathcal{D}(\boldsymbol{s}_i, \boldsymbol{s}_{gt}))}},
\label{eq:6}
\end{equation}

where $ \rm \mathcal{D}(\boldsymbol{s}_{i})$ of each predicted trajectory $\boldsymbol{s}_{i}$ is deﬁned by the L2 distance of its endpoint to ground truth endpoint, 
$\rm \mathcal{D}(\boldsymbol{s}_i, \boldsymbol{s}_{j}) = \left \| s_{i,T} - s_{j,T} \right \|_2 $.

\begin{equation}
    \rm \mathcal{L}_{conf} = \frac{1}{N}\sum_i^{N}D_{KL}( \lambda(\boldsymbol{s_i})|| \tau(\boldsymbol{y_i})).
\end{equation}

As we want the distribution of the predicted score to stay close to the target distribution calculated in Eq~\ref{eq:6}, We use the Kullback-Leibler Divergence as the loss function. 

\noindent\textbf{Classification Loss.} 
We introduces region classification loss to ensure model can identify correct region. When the proposal number $K$ is large, we find that using such auxiliary loss $\mathcal{L}_{cls}$ helps to regularize the confidence loss, which accelerates the convergence of mmTransformer + RTS. The details of $\mathcal{L}_{cls}$ is provided in appendix.

\noindent\textbf{Intermediate Layer Losses.} In order to accelerate the training process, we add Proposal Feature Decoder and the aforementioned combination of losses after each decoder layer of social constructor.

\noindent\textbf{Total Loss.} Since the total loss function can be viewed as the summation of multiple distinct tasks, we use multi-task learning approach in \cite{kendall2018multi,liebel2018auxiliary} to balance them.

\begin{equation}
        \rm
        \mathcal{L} = \frac{1}{\sigma_1^2} \mathcal{L}_{reg} + \frac{1}{\sigma_2^2} \mathcal{L}_{conf} + \frac{1}{\sigma_3^2} \mathcal{L}_{cls} + \sum_{i=1}^3 log(\sigma_i + 1),  
\end{equation}
where $\sigma_i, i\in \{1,2,3\}$ are learnable loss weights.

\subsection{Inference}
During inference, all the $K$ proposals are used to generate final results. In order to merge multimodal predictions, a selection algorithm inspired by the non-maximum suppression algorithm is used to reject near-duplicate trajectories based on the euclidean distances of endpoints (the detailed procedure can be found in appendix).

\section{Experiments}
\label{sec:exp}
\begin{figure*}[htb]
    \centering
    \includegraphics[width=0.8\linewidth]{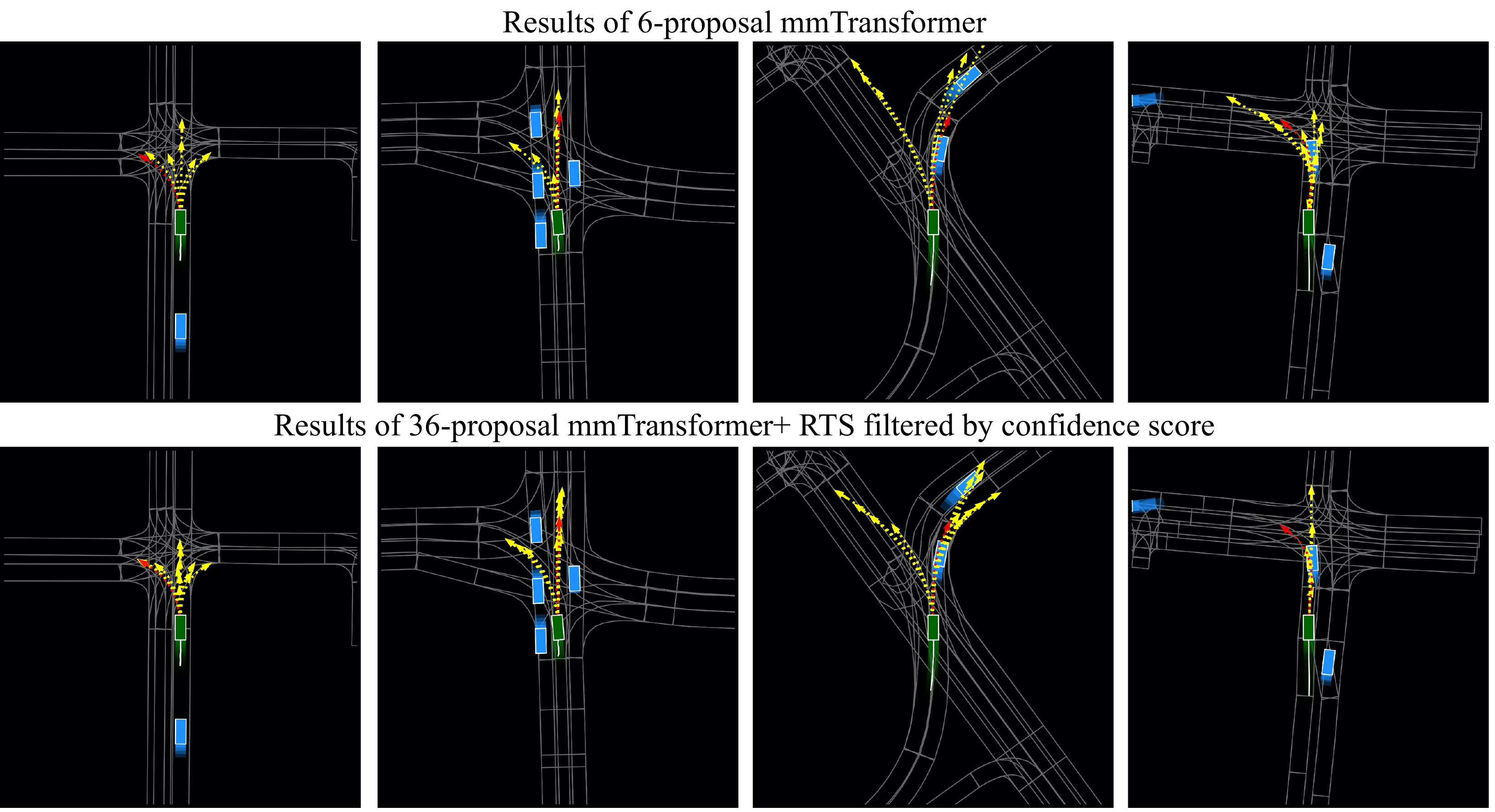}
    \caption{Qualitative comparison between mmTransformer (6 proposals) and mmTransformer+RTS (36 proposals) on four driving scenarios from Argoverse validation set. Green rectangle represents the target vehicle. The red and yellow arrows indicate the groundtruth and predicted trajectories with high confidence scores. For a clear visualization, we filter the trajectories with scores lower than the uniform probability in each scenario. It can be seen that most of plausible areas are covered by our prediction results. 
    }
    \label{fig:qualitatively_visualization}
\end{figure*}

\subsection{Experimental Setup}
\noindent\textbf{Dataset.} 
We perform experiments on Argoverse motion forecasting benchmark \cite{chang2019argoverse}, which involves 340k 5s long trajectory sequences and corresponding contextual information. The sequences are split into 205,942 training, 39,472 validation and 78,143 testing cases, respectively. Given a 2-second history trajectory and the context as inputs, the goal is to forecast the future movements of the target vehicle over the next 3 seconds. For each scenario, local map information can be represented as a set of centerline-based polylines from HD map. Besides, past trajectories and locations of the adjacent vehicles and the ego-car are also included in order to model the interaction between them.

\noindent\textbf{Metrics.}
We evaluate our model in terms of the widely used Average Displacement Error (ADE) and Final Displacement Error (FDE). Due to the multimodal nature of trajectory prediction, minADE, minFDE and miss rate (MR) of the top K (K=6) trajectories are also reported following the evaluation criteria of the Argoverse benchmark. 
\begin{table}[]
\begin{center}
\begin{tabular}{@{}c||ccccc@{}}
\hline
Methods                                         & minADE          & minFDE        & MR(\%)                \\ \hline \hline
NN ~\cite{chang2019argoverse}          & 1.7129          & 3.2870        & 53.69              \\
LSTM ED ~\cite{chang2019argoverse}     & 2.34            & 5.44          & -              \\
TNT  \cite{zhao2020tnt}  ($4^{th}$)             & 0.9358          & 1.5384          & 13.28             \\
LGCN \cite{liang2020learning}  ($7^{th}$)       & 0.8679          & 1.3640          & 16.34             \\
LA \cite{pan2019lane} ($21^{th}$)   & 0.9436          & 1.5486          & 21.79             \\
WIMF \cite{khandelwal2020if} ($8^{th}$)         & 0.8995          & 1.4220          & 16.69             \\
\hline
mmTrans. ($5^{th}$)                & \textbf{0.8435} & \textbf{1.3383} & 15.42             \\
mmTrans.+RTS ($1^{st}$)                & 0.8704          & 1.3688          & \textbf{13.00}     \\ 
\hline
\end{tabular}

\end{center}
\caption{Comparison with state-of-the-art methods on the Argoverse test set (K=6). Here, mmTrans. stands for 6-proposal mmTransformer, while mmTrans.+RTS stands for 36-proposal mmTransformer trained with RTS.}
\label{tab:compare_with_state_of_the_art}
\end{table}

\subsection{Results}
\label{sec:exp_comparison}
We compare our model with the state-of-the-art methods in the test set of Argoverse. The scores of different methods in Table~\ref{tab:compare_with_state_of_the_art} were extracted before the CVPR submission deadline (16/11/2020) from the \href{https://eval.ai/web/challenges/challenge-page/454/leaderboard/1279}{Argoverse Leaderboard}. 

As shown in Table~\ref{tab:compare_with_state_of_the_art}, we include the results of vanilla mmTransformer and mmTransformer model trained with RTS.
It shows that our models achieve the best performance in terms of all the metrics, which indicates that our method is capable of learning high quality proposal features by employing stacked transformers, and achieving promising multimodal results by using the RTS. Compared to the 6-proposal mmTransformer without RTS , we observe a slightly drop of minADE and minFDE in the 36-proposal mmTransformer with RTS. 
It is the large number of proposals that leads to the drop of minADE and minFDE. We explain the reason using outcome of 36-proposal mmTransformer as an example: Limitted by the fixed number (6) of final outputs, we discard the redundant candidate proposals to retain the diversity(\ie, MR) during the post-processing. As a side effect, the number of selected proposals in GT region is decreased, which may hurt the accuracy, \ie minFDE and minADE. We regard this as a trade-off between accuracy and diversity.
Besides, the accruacy drop caused by post-processing is a common issue in machine learning community \cite{bodla2017soft}, especially when selecting final predictions from a large candidate set.
 
\noindent\textbf{Visualization of Multimodal Motion Prediction.}
In Fig~\ref{fig:qualitatively_visualization}, we showcase multimodal prediction results of mmTransformer on the Argoverse validation set. We can see that mmTransformer generates trajectories covering all the plausible modes in each driving scenario. Although mmTransformer itself is capable to make reasonable predictions (column 1,2) with trajectory proposals, it fails to achieve comparable performance as mmTransformer+RTS in more challenging scenarios (column 3,4), as the modalities may concentrate in a specific area.

\begin{table}
\setlength{\belowdisplayskip}{3pt}
\begin{center}
\setlength{\tabcolsep}{0.5mm}{
    \begin{tabular}{@{}ccccc|ccc@{}}
    \hline
    Motion      & Map        & Social       & RTS        &Proposal    & minADE          & minFDE          & MR(\%)  \\
    \hline
    \hline
    \checkmark &             &              &            & 6          & 0.915	        & 1.681	          & 23.3 \\
    \checkmark & \checkmark  &              &            & 6          & 0.794	        & 1.284           & 14.4 \\
    \checkmark &             & \checkmark   &            & 6          & 0.826	        & 1.418	          & 17.3 \\
    \checkmark & \checkmark  & \checkmark   &            & 6          & \textbf{0.713}           & \textbf{1.153}           & 10.6 \\
    \checkmark & \checkmark  & \checkmark   &            & 36         & 0.833           & 1.453           & 17.6 \\
    \checkmark & \checkmark  & \checkmark   & \checkmark & 36         & 0.721           & 1.211  & \textbf{9.2}   \\
    \hline
    \end{tabular}
}
\end{center}
\caption{Ablation study on the effectiveness of different components of mmTransformer on the Argoverse validation dataset. 
As shown in the last two rows of Table 2, same model without RTS shows a poorer performance when other condition remain the same.
}
\label{tab:important_of_each_part}
\end{table}

\subsection{Ablation Study}
\label{sec:exp_xxx}
We first conduct ablation study to analyze the importance of each component in our model. Then, we evaluate two different partition methods, K-means and manual partition. We finally measure impact of the number of region $M$ and the number of proposal in each region $N$ on mmTransformer+RTS.

\noindent\textbf{Importance of Each Transformer Module and RTS.} To analyze the importance of each component in mmTransformer, we compare the results of several models on Argoverse validation set. We consider the motion extractor in mmTransformer as the baseline model, and progressively add other transformer modules to aggregate contextual information, and utilize RTS to encourage multimodal prediction. Observations can be drawn from the experimental results shown in Table~\ref{tab:important_of_each_part}.

Firstly, all the structural components contribute to the performance of the framework. We observe that the MR is improved from 23.3\% to 10.6\% by applying all the transformer modules (row 1-4). With the contextual information being captured and incorporated by each module, the model gains more comprehensive understanding of the scenario. For example, the map information captured by map aggregator brings useful road features (e.g. layout of lane lines), which beneﬁts the overall performance (from 23.3\% to 14.4\%). Also, The model with map aggregator and social constructor further promotes the MR to 10.6\%. It is noted that we stack these modules hierarchically rather than following a parallel design because of the logical relationship between different contextual information.

Besides, RTS facilities the final results by increasing the number of proposals to encourage multimodal prediction. As shown in Table~\ref{tab:important_of_each_part}, the region-based training strategy boosts the performance by a large margin, from 17.6\% to 9.2\% in MR. We attribute the large improvement to the large amount of proposals used by the RTS. However, training large number of proposals with vanilla strategy can not result in the comparable performance, since the optimization of a single proposal compromises the others under this setting. The comparison results demonstrate that RTS helps to preserve the modality information.

\begin{table}[]
\setlength{\belowdisplayskip}{2pt}
\begin{center}
\begin{tabular}{c|ccc}
 \hline
Partition method     & minADE           & minFDE            & MR(\%)         \\
\hline
\hline
K-means              & \textbf{0.72}    & \textbf{1.21}     & 9.21           \\
Manual partition   & 0.73             & 1.23              & \textbf{9.13}   \\
\hline
\end{tabular}
\end{center}
\caption{Impact of different partition algorithms on the Argoverse validation dataset.}
\label{tab:partition methods}
\end{table}

\begin{figure*}[htb]
    \centering
    \includegraphics[width=0.8\linewidth]{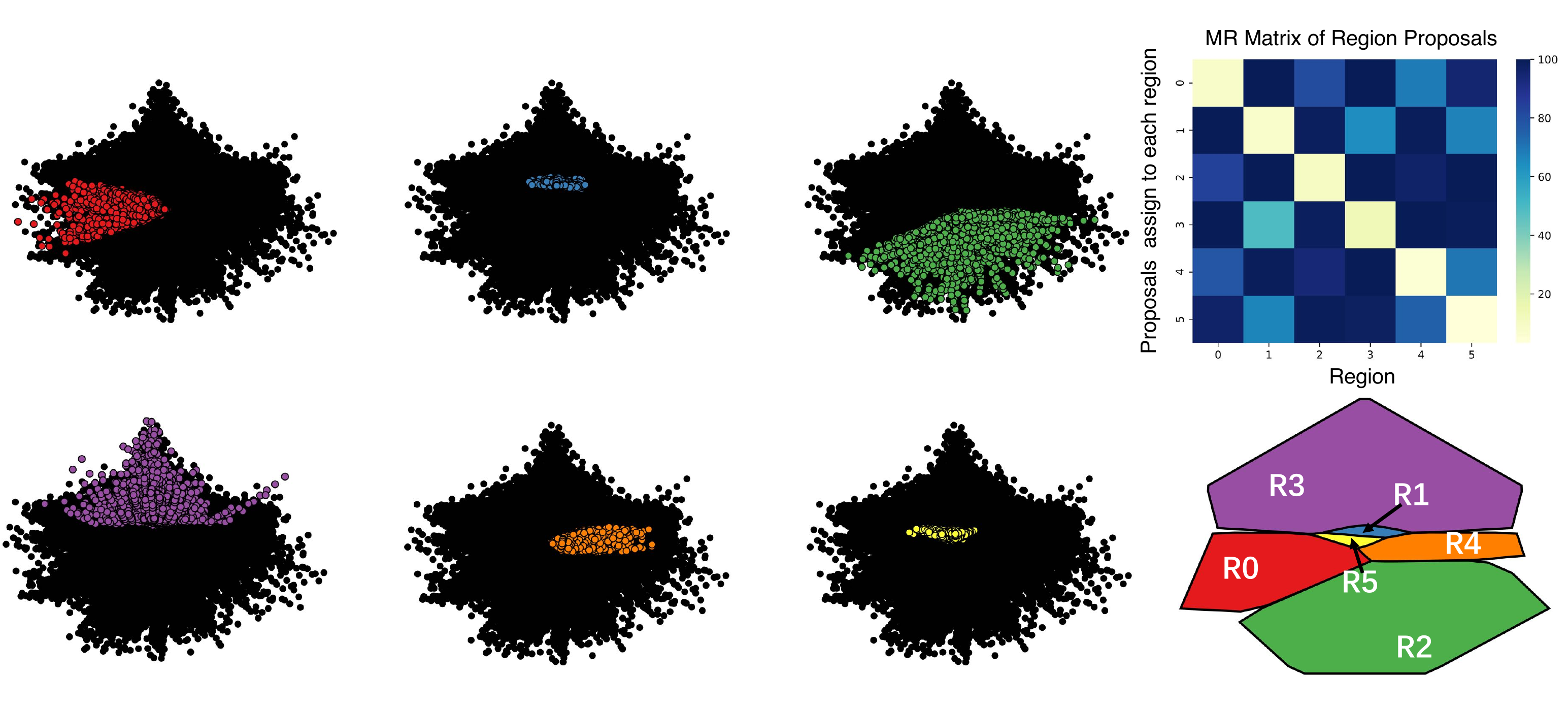}
    \caption{Visualization of the multimodal prediction results on Argoverse validation set. We utilize all trajectory proposals of mmTransformer to generate multiple trajectories for each scenario and visualize all the predicted endpoints in the figures (left three columns). 
    For clear illustration, we filter the points with confidences lower than the uniform probability($1/K$). The background represents all the predicted results and colored points indicate the prediction results of a specific group of proposals (regional proposals). We observe that the endpoints generated by each group of regional proposals are within the associated region. 
    Miss Rate (MR) matrix of regional proposals is shown on the upper right, where the value in each cell $(i,j)$ represents the MR calculated by proposals assigned to region $i$ and ground truth in region $j$. 
    The proposals possess high accuracy when the GT is located in their region. For reference, the layout of the regions produced by constrained K-means\cite{wagstaff2001constrained} is shown in the bottom right.}
    \label{fig:qualitatively_partition_visualization}
\end{figure*}
\noindent\textbf{Spatial partitions.} We evaluate two ways to divide the surrounding space for RTS. The first one adopts constrained K-means \cite{wagstaff2001constrained}, while we manually split the space into fan-shaped regions(similar with Fig~\ref{fig:region-based Training strategy}) for the another one. Training samples are evenly distributed in each region for data balance.
For a fair comparison, we partition the space into 6 regions, according to the number of regions, and assign 6 proposals (represented as regional proposals) to each region. 
Compared to K-means based partition algorithm, manual partition can successfully divide some blurry samples to correct region. Since we assume that the misclassified samples may perturb the learning of regional proposal, manual partition can, therefore, achieve slightly higher performance, as shown in Table~\ref{tab:partition methods}. How to classify the samples to help training remains to be an open topic.

\begin{table}[]
\setlength{\belowdisplayskip}{3pt}
\begin{center}
\begin{tabular}{c|ccc}
\diagbox{N}{MR(\%)}{M}  & 3         &6              & 9      \\
\hline
\hline
1                       & 28.87     & 20.67         & 23.86  \\ 
6                       & 11.65     & \textbf{9.21} & 9.37	 \\
8                       & 12.96     & 9.23          & 9.31   \\
\hline
\end{tabular}
\end{center}
\caption{Impact of number of region $M$ and the number of proposal in each region $N$ on the Argoverse validation dataset. }
\label{tab:region and proposal number}
\end{table}

\noindent\textbf{Number of Proposals.} 
We further conduct experiments to explore the appropriate number of proposals and regions utilized in mmTransformer+RTS. We hypothesis that the number of regions ($M$) and the number of proposals in each region ($N$) jointly control the concentration and coverage of predicted trajectories. We find that the ratio between $M$ and $N$ affects the performance significantly when total number of proposal is not very large. Specifically, the performance drops when the ratio is far way from 1.
However, the performance increases marginally and even becomes worse when the total number of proposal is large, regardless of the changing of ratio. According to our experiments, the model gives the most desirable performance when $M$ and $N$ both equal to 6.

\noindent\textbf{Visualization of Region-based Training Strategy.} Fig~\ref{fig:qualitatively_partition_visualization} illustrates the effectiveness of RTS (36 proposals). We conduct the experiments on Argoverse validation set. As shown in the MR matrix of Fig~\ref{fig:qualitatively_partition_visualization}, cell $(i,j)$ represents the missing rate of proposals assigned to i-th region (named as region proposals) in predicting all cases that belong to j-th region. The low MR in diagonal indicates that the regional proposals have learned specialization in assigned region. We observe that each proposal tends to generate the trajectory which ends in the preassigned region, which demonstrates that mmTransformer has learned different modalities in a region-based manner.

\section{Conclusion}
We develop a transformer-based motion prediction model called mmTransformer for accurate multimodal prediction. A novel partition training method is introduced to improve the multimodal prediction. The experiments show the competitive result on the Argoverse benchmark.

\noindent\textbf{Acknowledgement:} This project was partially supported by the Centre for Perceptual and Interactive Intelligence (CPII) Ltd under the Innovation and Technology Fund. The authors would like to thank Xin Zhang for his insightful discussions.

\section*{Appendix}
\begin{appendix}
\section{Implementation Details}
Following previous transformer-based approaches \cite{carion2020end,vaswani2017attention}, we utilize AdamW\cite{loshchilov2017decoupled} as the optimizer, with the initial learning rate, weight decay and gradient max norm set to $1\times10^{-3}$, $1\times10^{-4}$ and $0.1$ respectively. All parameters in mmTransformer are initialized using Xavier initialization \cite{glorot2010understanding}. All transformer modules in mmTransformer contain 128 hidden units. Each transformer module has two encoder layers and two decoder layers, except for the decoder of social constructor, which contains four layers. The map information in our implementation covers the $65m\times65m$ local region, centered at the position of target agent at the last observation point. Besides, the heading direction of the target agent at last observation point is aligned to $+y$ axis, as mentioned before. To enhance the robustness of the model, we further conduct data augmentation by flipping the trajectories horizontally and randomly masking the trajectories at the first ten time steps.

\begin{figure}
    \centering
    \includegraphics[width=\linewidth]{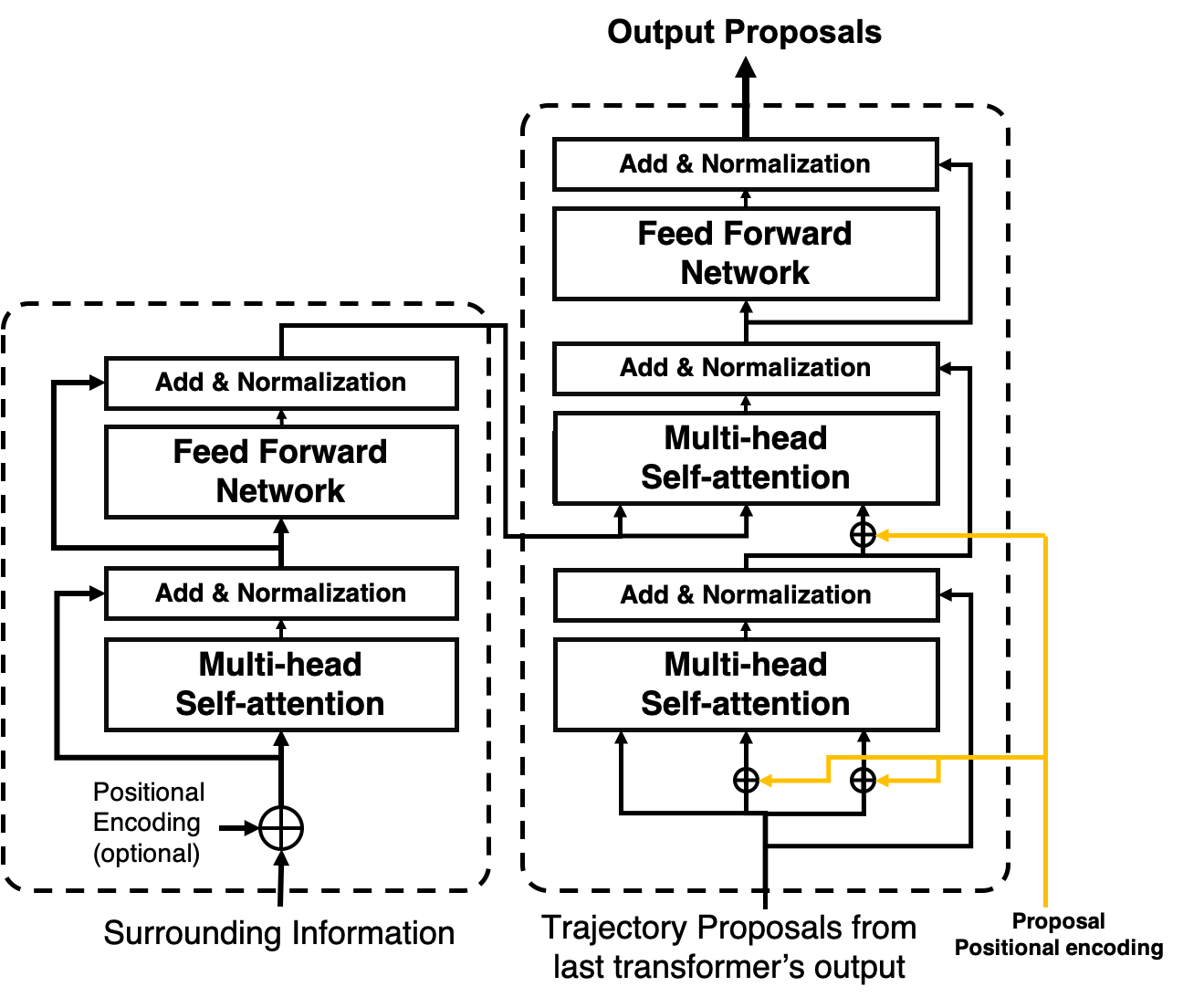}
    \caption{The detailed structure of the transformer module.}
    \label{fig:detailed_structure}
\end{figure}

\section{Detailed Architecture}
The detailed architecture of the transformer used in mmTransformer is visualized in Fig~\ref{fig:detailed_structure}. Surrounding information is passed to the encoder to derive memory of the contextual features. Besides, spatial positional encodings are added to the queries and keys at each multi-head self-attention layer. Then, the decoder receives proposals (randomly initialized), positional encoding of proposals, as well as encoder memory, and produces the refined trajectory proposals through multiple multi-head self-attention and decoder-encoder attention layers. It is noted that the ﬁrst self-attention layer in the ﬁrst decoder layer of motion extractor can be skipped.

\section{Classification Loss}
The idea is to encourage the trajectory proposals which are assigned to the ground truth region to have higher scores. Specifically, each logit in loss term is the sum of the scores belonging to the corresponding region 
\begin{equation}
    P = \{p_i \in \mathbb{R}: p_i =\sum_{j=i}^{i+N} c_{j} \},
\end{equation}
where $i \in \{1, N+1, 2N+1,\dots,(M-1)N+1 \}$, then we apply the cross entropy loss to calculate the penalty as
\begin{equation}
    \rm
    \mathcal{L}_{cls} = -\sum_{i=1}^{M}\delta(i-gt)\log p_i,  
\end{equation}
where the $\rm gt$ is the ground truth region index and $\delta$ is a indicator function. The auxiliary loss benefits to the generalization and convergence of our model.

\begin{figure}
    \centering
    \includegraphics[width =\linewidth]{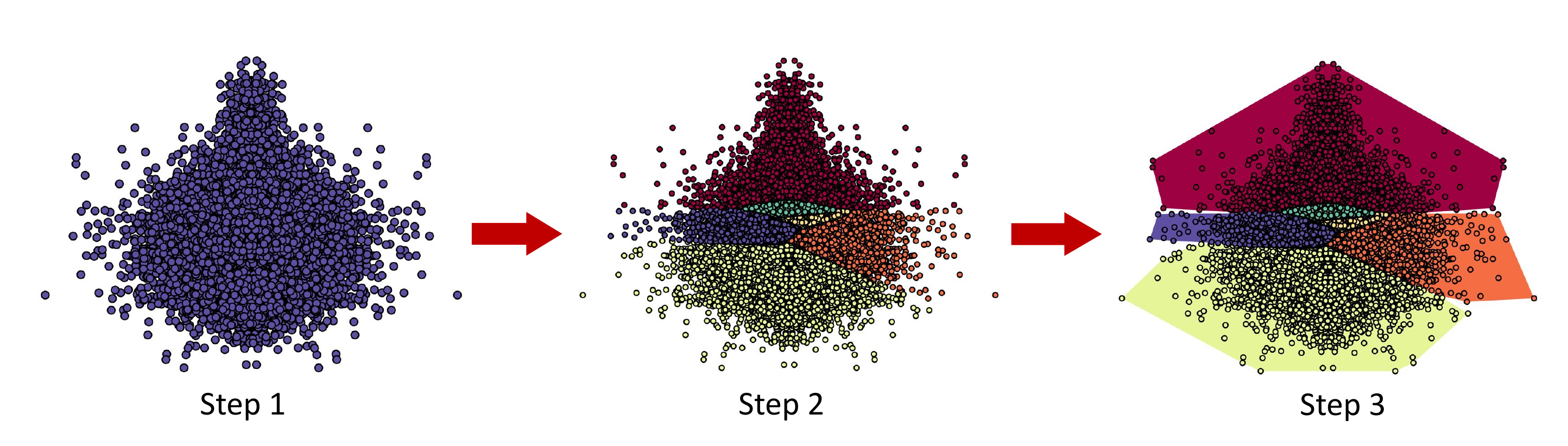}
    \caption{Visualization of the partition procedure.}
    \label{fig:partition_vis}
\end{figure}

\section{The Procedure of Partition Algorithm}
Let's take K-means as a example to described partition procedure. Step1: Extract all normalized GT trajectory endpoints, using normalization described in line 435 of our paper. Step2: Apply constrained K-means~\cite{wagstaff2001constrained} to divide these samples into $M$ clusters equally. Step3: Find the vertices of each region with convex hull algorithm; gather these vertices to form the regions. The procedure is visualized in Fig~\ref{fig:partition_vis}

\section{Inference}
During the inference stage, we utilize NMS algorithm to filter duplicated trajectories.
The detail of NMS algorithm goes as follow: we first sort the predicted trajectories according to their confidence scores in descending order, and then pick them greedily. Specifically, we set a threshold and exclude trajectories that are close to any of the selected trajectories. We keep repeating above two steps until collecting sufficient predicted trajectories.

\end{appendix}

{\small
\bibliographystyle{ieee_fullname}
\bibliography{Arxiv_version}

\begin{thebibliography}{10}\itemsep=-1pt

\bibitem{alahi2016social}
Alexandre Alahi, Kratarth Goel, Vignesh Ramanathan, Alexandre Robicquet, Li
  Fei-Fei, and Silvio Savarese.
\newblock Social lstm: Human trajectory prediction in crowded spaces.
\newblock In {\em Proceedings of the IEEE conference on computer vision and
  pattern recognition}, pages 961--971, 2016.

\bibitem{altche2017lstm}
Florent Altch{\'e} and Arnaud de La~Fortelle.
\newblock An lstm network for highway trajectory prediction.
\newblock In {\em 2017 IEEE 20th International Conference on Intelligent
  Transportation Systems (ITSC)}, pages 353--359. IEEE, 2017.

\bibitem{bodla2017soft}
Navaneeth Bodla, Bharat Singh, Rama Chellappa, and Larry~S Davis.
\newblock Soft-nms--improving object detection with one line of code.
\newblock In {\em Proceedings of the IEEE international conference on computer
  vision}, pages 5561--5569, 2017.

\bibitem{carion2020end}
Nicolas Carion, Francisco Massa, Gabriel Synnaeve, Nicolas Usunier, Alexander
  Kirillov, and Sergey Zagoruyko.
\newblock End-to-end object detection with transformers.
\newblock {\em arXiv preprint arXiv:2005.12872}, 2020.

\bibitem{chai2019multipath}
Yuning Chai, Benjamin Sapp, Mayank Bansal, and Dragomir Anguelov.
\newblock Multipath: Multiple probabilistic anchor trajectory hypotheses for
  behavior prediction.
\newblock {\em arXiv preprint arXiv:1910.05449}, 2019.

\bibitem{chang2019argoverse}
Ming-Fang Chang, John Lambert, Patsorn Sangkloy, Jagjeet Singh, Slawomir Bak,
  Andrew Hartnett, De Wang, Peter Carr, Simon Lucey, Deva Ramanan, et~al.
\newblock Argoverse: 3d tracking and forecasting with rich maps.
\newblock In {\em Proceedings of the IEEE Conference on Computer Vision and
  Pattern Recognition}, pages 8748--8757, 2019.

\bibitem{chu2019neural}
Hang Chu, Daiqing Li, David Acuna, Amlan Kar, Maria Shugrina, Xinkai Wei,
  Ming-Yu Liu, Antonio Torralba, and Sanja Fidler.
\newblock Neural turtle graphics for modeling city road layouts.
\newblock In {\em Proceedings of the IEEE International Conference on Computer
  Vision}, pages 4522--4530, 2019.

\bibitem{deo2018convolutional}
Nachiket Deo and Mohan~M Trivedi.
\newblock Convolutional social pooling for vehicle trajectory prediction.
\newblock In {\em Proceedings of the IEEE Conference on Computer Vision and
  Pattern Recognition Workshops}, pages 1468--1476, 2018.

\bibitem{devlin2018bert}
Jacob Devlin, Ming-Wei Chang, Kenton Lee, and Kristina Toutanova.
\newblock Bert: Pre-training of deep bidirectional transformers for language
  understanding.
\newblock {\em arXiv preprint arXiv:1810.04805}, 2018.

\bibitem{djuric2018motion}
Nemanja Djuric, Vladan Radosavljevic, Henggang Cui, Thi Nguyen, Fang-Chieh
  Chou, Tsung-Han Lin, and Jeff Schneider.
\newblock Motion prediction of traffic actors for autonomous driving using deep
  convolutional networks.
\newblock {\em arXiv preprint arXiv:1808.05819}, 2018.

\bibitem{fang2020tpnet}
Liangji Fang, Qinhong Jiang, Jianping Shi, and Bolei Zhou.
\newblock Tpnet: Trajectory proposal network for motion prediction.
\newblock In {\em Proceedings of the IEEE/CVF Conference on Computer Vision and
  Pattern Recognition}, pages 6797--6806, 2020.

\bibitem{gao2020vectornet}
Jiyang Gao, Chen Sun, Hang Zhao, Yi Shen, Dragomir Anguelov, Congcong Li, and
  Cordelia Schmid.
\newblock Vectornet: Encoding hd maps and agent dynamics from vectorized
  representation.
\newblock In {\em Proceedings of the IEEE/CVF Conference on Computer Vision and
  Pattern Recognition}, pages 11525--11533, 2020.

\bibitem{giuliari2020transformer}
Francesco Giuliari, Irtiza Hasan, Marco Cristani, and Fabio Galasso.
\newblock Transformer networks for trajectory forecasting.
\newblock {\em arXiv preprint arXiv:2003.08111}, 2020.

\bibitem{glorot2010understanding}
Xavier Glorot and Yoshua Bengio.
\newblock Understanding the difficulty of training deep feedforward neural
  networks.
\newblock In {\em Proceedings of the thirteenth international conference on
  artificial intelligence and statistics}, pages 249--256, 2010.

\bibitem{sgan}
Agrim Gupta, Justin Johnson, Li Fei-Fei, Silvio Savarese, and Alexandre Alahi.
\newblock Social gan: Socially acceptable trajectories with generative
  adversarial networks.
\newblock In {\em Proceedings of the IEEE Conference on Computer Vision and
  Pattern Recognition}, pages 2255--2264, 2018.

\bibitem{hochreiter1997long}
Sepp Hochreiter and J{\"u}rgen Schmidhuber.
\newblock Long short-term memory.
\newblock {\em Neural computation}, 9(8):1735--1780, 1997.

\bibitem{hong2019rules}
Joey Hong, Benjamin Sapp, and James Philbin.
\newblock Rules of the road: Predicting driving behavior with a convolutional
  model of semantic interactions.
\newblock In {\em Proceedings of the IEEE Conference on Computer Vision and
  Pattern Recognition}, pages 8454--8462, 2019.

\bibitem{kendall2018multi}
Alex Kendall, Yarin Gal, and Roberto Cipolla.
\newblock Multi-task learning using uncertainty to weigh losses for scene
  geometry and semantics.
\newblock In {\em Proceedings of the IEEE conference on computer vision and
  pattern recognition}, pages 7482--7491, 2018.

\bibitem{khandelwal2020if}
Siddhesh Khandelwal, William Qi, Jagjeet Singh, Andrew Hartnett, and Deva
  Ramanan.
\newblock What-if motion prediction for autonomous driving.
\newblock {\em arXiv preprint arXiv:2008.10587}, 2020.

\bibitem{lee2017desire}
Namhoon Lee, Wongun Choi, Paul Vernaza, Christopher~B Choy, Philip~HS Torr, and
  Manmohan Chandraker.
\newblock Desire: Distant future prediction in dynamic scenes with interacting
  agents.
\newblock In {\em Proceedings of the IEEE Conference on Computer Vision and
  Pattern Recognition}, pages 336--345, 2017.

\bibitem{li2020end}
Lingyun~Luke Li, Bin Yang, Ming Liang, Wenyuan Zeng, Mengye Ren, Sean Segal,
  and Raquel Urtasun.
\newblock End-to-end contextual perception and prediction with interaction
  transformer.
\newblock {\em arXiv preprint arXiv:2008.05927}, 2020.

\bibitem{liang2020learning}
Ming Liang, Bin Yang, Rui Hu, Yun Chen, Renjie Liao, Song Feng, and Raquel
  Urtasun.
\newblock Learning lane graph representations for motion forecasting.
\newblock In {\em European Conference on Computer Vision}, pages 541--556.
  Springer, 2020.

\bibitem{liebel2018auxiliary}
Lukas Liebel and Marco K{\"o}rner.
\newblock Auxiliary tasks in multi-task learning.
\newblock {\em arXiv preprint arXiv:1805.06334}, 2018.

\bibitem{loshchilov2017decoupled}
Ilya Loshchilov and Frank Hutter.
\newblock Decoupled weight decay regularization.
\newblock {\em arXiv preprint arXiv:1711.05101}, 2017.

\bibitem{luscher2019rwth}
Christoph L{\"u}scher, Eugen Beck, Kazuki Irie, Markus Kitza, Wilfried Michel,
  Albert Zeyer, Ralf Schl{\"u}ter, and Hermann Ney.
\newblock Rwth asr systems for librispeech: Hybrid vs attention--w/o data
  augmentation.
\newblock {\em arXiv preprint arXiv:1905.03072}, 2019.

\bibitem{mercat2020multi}
Jean Mercat, Thomas Gilles, Nicole El~Zoghby, Guillaume Sandou, Dominique
  Beauvois, and Guillermo~Pita Gil.
\newblock Multi-head attention for multi-modal joint vehicle motion
  forecasting.
\newblock In {\em 2020 IEEE International Conference on Robotics and Automation
  (ICRA)}, pages 9638--9644. IEEE, 2020.

\bibitem{pan2019lane}
Jiacheng Pan, Hongyi Sun, Kecheng Xu, Yifei Jiang, Xiangquan Xiao, Jiangtao Hu,
  and Jinghao Miao.
\newblock Lane attention: Predicting vehicles' moving trajectories by learning
  their attention over lanes.
\newblock {\em arXiv preprint arXiv:1909.13377}, 2019.

\bibitem{phan2020covernet}
Tung Phan-Minh, Elena~Corina Grigore, Freddy~A Boulton, Oscar Beijbom, and
  Eric~M Wolff.
\newblock Covernet: Multimodal behavior prediction using trajectory sets.
\newblock In {\em Proceedings of the IEEE/CVF Conference on Computer Vision and
  Pattern Recognition}, pages 14074--14083, 2020.

\bibitem{sadeghian2019sophie}
Amir Sadeghian, Vineet Kosaraju, Ali Sadeghian, Noriaki Hirose, Hamid
  Rezatofighi, and Silvio Savarese.
\newblock Sophie: An attentive gan for predicting paths compliant to social and
  physical constraints.
\newblock In {\em Proceedings of the IEEE Conference on Computer Vision and
  Pattern Recognition}, pages 1349--1358, 2019.

\bibitem{tang2019multiple}
Charlie Tang and Russ~R Salakhutdinov.
\newblock Multiple futures prediction.
\newblock In {\em Advances in Neural Information Processing Systems}, pages
  15424--15434, 2019.

\bibitem{vaswani2017attention}
Ashish Vaswani, Noam Shazeer, Niki Parmar, Jakob Uszkoreit, Llion Jones,
  Aidan~N Gomez, {\L}ukasz Kaiser, and Illia Polosukhin.
\newblock Attention is all you need.
\newblock In {\em Advances in neural information processing systems}, pages
  5998--6008, 2017.

\bibitem{wagstaff2001constrained}
Kiri Wagstaff, Claire Cardie, Seth Rogers, Stefan Schr{\"o}dl, et~al.
\newblock Constrained k-means clustering with background knowledge.
\newblock In {\em Icml}, volume~1, pages 577--584, 2001.

\bibitem{yagi2018future}
Takuma Yagi, Karttikeya Mangalam, Ryo Yonetani, and Yoichi Sato.
\newblock Future person localization in first-person videos.
\newblock In {\em Proceedings of the IEEE Conference on Computer Vision and
  Pattern Recognition}, pages 7593--7602, 2018.

\bibitem{zhang2020novel}
Yifan Zhang, Jinghuai Zhang, Jindi Zhang, Jianping Wang, Kejie Lu, and Jeff
  Hong.
\newblock A novel learning framework for sampling-based motion planning in
  autonomous driving.
\newblock In {\em Proceedings of the AAAI Conference on Artificial
  Intelligence}, volume~34, pages 1202--1209, 2020.

\bibitem{zhao2020tnt}
Hang Zhao, Jiyang Gao, Tian Lan, Chen Sun, Benjamin Sapp, Balakrishnan
  Varadarajan, Yue Shen, Yi Shen, Yuning Chai, Cordelia Schmid, et~al.
\newblock Tnt: Target-driven trajectory prediction.
\newblock {\em arXiv preprint arXiv:2008.08294}, 2020.

\end{thebibliography}
}

\end{document}